\documentclass[journal]{IEEEtran}

\usepackage{graphicx}  
\usepackage{float}  
\usepackage{amsthm,amsmath,amssymb}
\usepackage{mathrsfs}
\usepackage{dutchcal}
\usepackage{multirow}
\usepackage{multicol}
\usepackage{setspace}
\usepackage{caption}
\usepackage{cite}
\usepackage{color}
\usepackage[colorlinks,linkcolor=blue,citecolor=blue]{hyperref}
\usepackage{makecell}
\usepackage{array}
\usepackage{booktabs}
\usepackage{colortbl}
\usepackage{color}
\usepackage{graphicx}
\usepackage{subfig}

\newcommand{\etal}{\textit{et al}.}
\newcommand{\ie}{\textit{i}.\textit{e}.}
\newcommand{\eg}{\textit{e}.\textit{g}.}
\newcommand{\etc}{\textit{etc}}

\begin{document}
%
\title{A Weakly Supervised Learning Framework for Salient Object Detection via Hybrid Labels}
%
%
%

\author{Runming Cong,~\IEEEmembership{Member,~IEEE,}
        Qi Qin,
        Chen Zhang,
        Qiuping Jiang,
        Shiqi Wang,\\
        Yao Zhao,~\IEEEmembership{Senior Member,~IEEE,}
        and~Sam~Kwong,~\IEEEmembership{Fellow,~IEEE}

\thanks{Runmin Cong is with the Institute of Information Science, Beijing Jiaotong University, Beijing 100044, China, also with the Beijing Key Laboratory of Advanced Information Science and Network Technology, Beijing 100044, China, and also with the Department of Computer Science, City University of Hong Kong, Hong Kong SAR, China (e-mail: rmcong@bjtu.edu.cn).}
\thanks{Qi Qin, Chen Zhang, and Yao Zhao are with the Institute of Information Science, Beijing Jiaotong University, Beijing 100044, China, and also with the Beijing Key Laboratory of Advanced Information Science and Network Technology, Beijing 100044, China (e-mail: qiqin96@bjtu.edu.cn;chen.zhang@bjtu.edu.cn; yzhao@bjtu.edu.cn).}
\thanks{Qiuping Jiang is with the School of Information Science and Engineering, Ningbo University, Ningbo 315211, China (e-mail: jiangqiuping@nbu.edu.cn).}
\thanks{Shiqi Wang and Sam Kwong are with the Department of Computer Science, City University of Hong Kong, Hong Kong SAR, China, and also with the City University of Hong Kong Shenzhen Research Institute, Shenzhen 51800, China (e-mail: shiqwang@cityu.edu.hk; cssamk@cityu.edu.hk).}
\thanks{Copyright \copyright 2022 IEEE. Personal use of this material is permitted. However, permission to use this material for any other purposes must be obtained from the IEEE by sending an email to pubs-permissions@ieee.org.}
}

%
%

\markboth{IEEE Transactions on Circuits and Systems for Video Technology}%
{Shell \MakeLowercase{\textit{et al.}}: Bare Demo of IEEEtran.cls for IEEE Journals}
%



\maketitle

\begin{abstract}
Fully-supervised salient object detection (SOD) methods have made great progress, but such methods often rely on a large number of pixel-level annotations, which are time-consuming and labour-intensive. In this paper, we focus on a new weakly-supervised SOD task under hybrid labels, where the supervision labels include a large number of coarse labels generated by the traditional unsupervised method and a small number of real labels. To address the issues of label noise and quantity imbalance in this task, we design a new pipeline framework with three sophisticated training strategies. In terms of model framework, we decouple the task into label refinement sub-task and salient object detection sub-task, which cooperate with each other and train alternately. Specifically, the R-Net is designed as a two-stream encoder-decoder model equipped with Blender with Guidance and Aggregation Mechanisms (BGA), aiming to rectify the coarse labels for more reliable pseudo-labels, while the S-Net is a replaceable SOD network supervised by the pseudo labels generated by the current R-Net. Note that, we only need to use the trained S-Net for testing.
Moreover, in order to guarantee the effectiveness and efficiency of network training, we design three training strategies, including alternate iteration mechanism, group-wise incremental mechanism, and credibility verification mechanism. Experiments on five SOD benchmarks show that our method achieves competitive performance against weakly-supervised/unsupervised methods both qualitatively and quantitatively.
The code and results can be found from the link of \url{https://rmcong.github.io/proj\_Hybrid-Label-SOD.html}.
\end{abstract}


%
\IEEEpeerreviewmaketitle

\section{Introduction}
%
%
%
%
Salient object detection (SOD) task aims to locate the most attractive and interesting objects or regions from an image, which is consistent with the human visual attention mechanism, and has been applied to image segmentation \cite{sun2019saliency, DBLP:journals/tcsv/ShiXZZWLZ22, DBLP:journals/tcsv/JiSLCM21,crmACMMM20-1,BCNet,crmcovid1,crmcovid2}, object tracking \cite{li2019siamrpn++, DBLP:journals/tcsv/LinFHGT21}, image enhancement \cite{crmbrain,crmJEI,crmCVPR21,crmSPIC,crmsrijcai,crmSRInpaintor,crmdsr2019tip,crmblindSR22}, and other vision tasks.
In recent years, fully-supervised SOD models based on deep learning have made great breakthroughs in performance \cite{chen2020global,crmCoADNet,crmDPANet,crmDBLP:journals/tip/WenYZCSZZBD21,crm2020tc,crmDBLP:journals/tmm/MaoJCGSK22,crmglnet,crmMM21}, but these models usually require a large number of pixel-level labels for training, while such labeling costs are obviously very expensive. Therefore, weakly-supervised or unsupervised SOD methods have received increasing attention from both academia and industry, aiming to reduce or get rid of the reliance on the elaborately labeled data. Some related areas such as weakly supervised semantic segmentation \cite{9440699,DBLP:journals/pr/ZhangXWHLZ22,DBLP:journals/corr/abs-2108-01296,DBLP:conf/aaai/ZhangXWSH20,zhangijcai2022},
light field SOD \cite{DBLP:journals/corr/abs-2204-13456,crm-acmmm}, remote sensing SOD \cite{DBLP:journals/corr/abs-2202-03501,crmRRNet,crm2019tgrs,crm-nc,dafnet,crm2022rsi}, and visual grounding \cite{9667277,DBLP:journals/pami/SunXLLG21} have also been developed.

\begin{figure}[!t]

  \subfloat[RGB Image]{\includegraphics[width=0.3\linewidth]{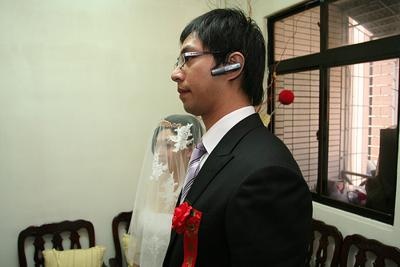}}
 \hfill 	
  \subfloat[Scribble Label]{\includegraphics[width=0.3\linewidth]{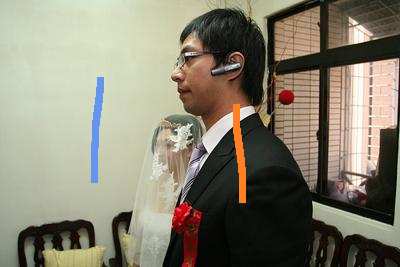}}
 \hfill	
  \subfloat[Point Label]{\includegraphics[width=0.3\linewidth]{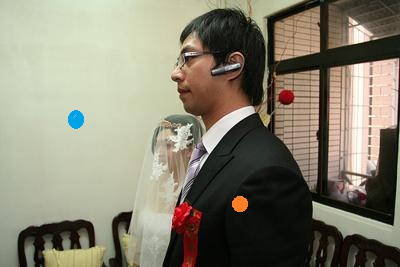}}
  \newline
  \subfloat[Pixel-level Label]{\includegraphics[width=0.3\linewidth]{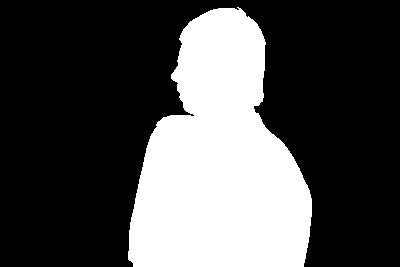}}
 \hfill 	
  \subfloat[Image-level Label]{\includegraphics[width=0.3\linewidth]{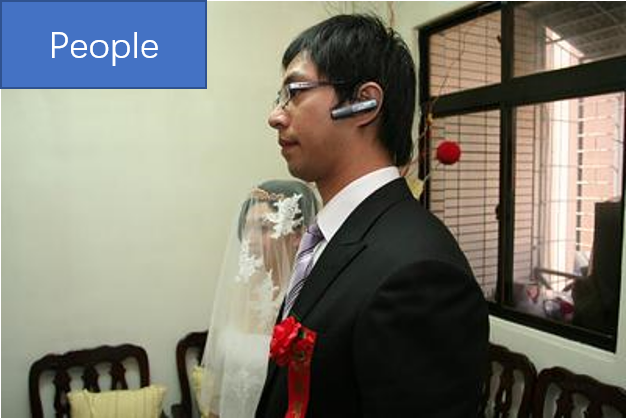}}
 \hfill	
  \subfloat[Coarse Label]{\includegraphics[width=0.3\linewidth]{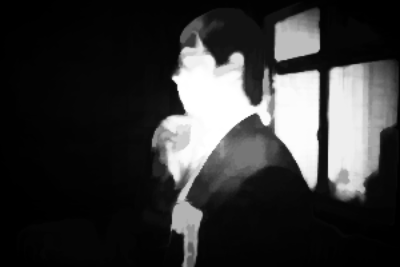}}
\caption{Several types of saliency supervision. (a) Original RGB image; (b) Weakly-supervised scribble label; (c) Weakly-supervised point label; (d) Pixel-level label; (e) Weakly-supervised image-level label; (f) Unsupervised coarse label.}
\label{level}
\end{figure}

According to the given labeled data, weakly-supervised/unsupervised SOD methods can be roughly divided into the following categories:
(1) Weakly-supervised scribble label supervision as shown in Fig. \ref{level}(b), that is, the parts of the foreground and background regions of each training sample are outlined in a scribble way.
\begin{figure}[!t]
    \centering
    \includegraphics[scale=0.28]{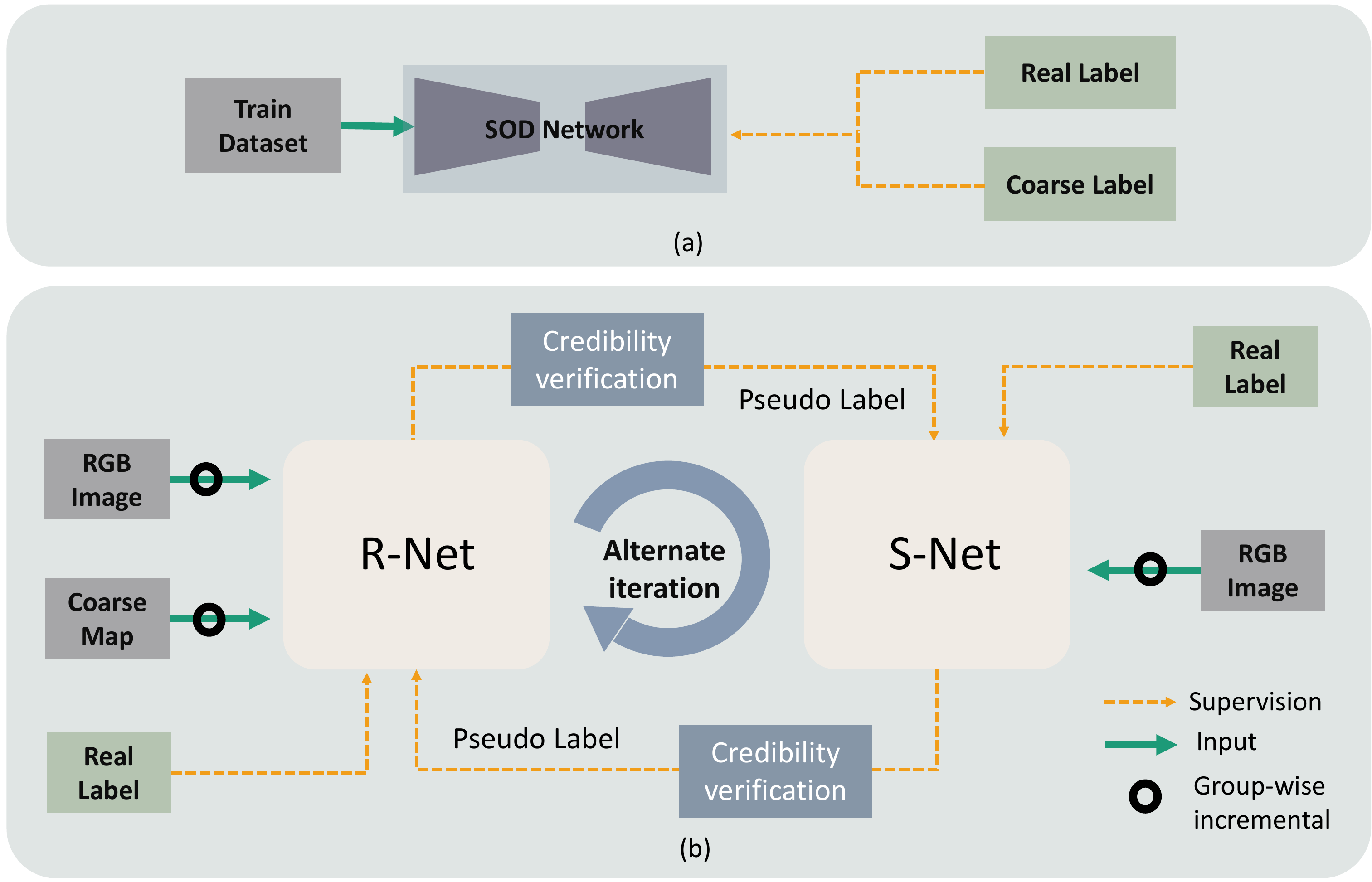}
    \caption{(a) A simple solution for training the SOD model with coarse and real labels. (b) The proposed alternate learning framework for weakly-supervised SOD task under the hybrid label, consisting of a Refine Network (R-Net) and a Saliency Network (S-Net). These two networks cooperate with each other and train alternately. During training, both networks employ a group-wise incremental mechanism to address the imbalance between real-labeled data and pseudo-labeled data, and use a credibility verification mechanism to ensure that the two networks can provide credible labels.
    }
    \label{fig:general-framework}
\end{figure}
(2) Weakly-supervised point label supervision as shown in Fig. \ref{level}(c), that is, the foreground and background regions of each training sample are marked with only one point respectively.
(3) Weakly-supervised image-level label supervision as shown in Fig. \ref{level}(e), that is, only the category of salient object is known during training.
(4) Unsupervised coarse label supervision as shown in Fig. \ref{level}(f), that is, the saliency map generated by the existing unsupervised traditional methods is used as the label of the training sample.
Generally speaking, the weaker the supervision information, the more limited the detection performance.
Compared to image-level labels and sparse labels (\ie, scribble and point labels) as supervision information, the information given by coarse label is uncontrollable and may introduce some inevitable distracting noise, so it provides weaker supervision than others.
Different from the above-mentioned forms of supervision, we construct a new supervision form for the first time to solve the weakly-supervised SOD task by releasing part of the real labels on the basis of unsupervised labels, called hybrid labels.
The hybrid label supervision consists of two parts, that is, a small number of pixel-level real labels and a large number of coarse labels generated by the existing unsupervised SOD models (\eg, one tenth of the fully-supervised pixel-wise annotations).
Such weakly supervision form is expected to achieve better detection performance at a smaller annotation cost.
But with this kind of supervision, the weakly-supervised SOD task becomes more challenging due to the unreliability of coarse labels and the imbalance between real-labeled and coarse-labeled data.
Specifically, on the one hand, the coarse labels are generated by traditional unsupervised methods and necessarily contain a lot of noise and mislabeling.
If the network is trained with such labels all the time, the network will gradually become chaotic and disabled, seriously affecting the final performance.
On the other hand, the proportion of real labels and coarse labels in the hybrid label setting is severely imbalanced (\eg, 1:9), and the network learning will collapse if the mixed training is directly performed. Therefore, in order to address these issues, we propose a new weakly-supervised SOD framework with hybrid labels from the perspective of pipeline structure and training strategy.


For such a new weakly-supervised learning task set in this paper, the key problems we need to solve are also different from the existing methods, and thus our model framework and technical implementation are also different.
To address the problem of unreliable coarse label and imbalanced sample size under this new hybrid supervision, different from the previous single-stage SOD framework \cite{wang2017learning,zhang2018deep,DBLP:journals/tcsv/ZhengTZML21}, we decouple the weakly-supervised SOD task into two sub-tasks of coarse label refinement and salient object detection, and construct a joint learning framework as shown in Fig. \ref{fig:general-framework}(b), consisting of a Refinement Network (R-Net) and a Saliency Network (S-Net). These two networks cooperate with each other and train alternately.
To achieve the R-Net, a two-stream encoder-decoder model equipped with Blender with Guidance and Aggregation Mechanisms (BGA) is designed for coarse label refinement, including a saliency-refinement mainstream branch and an RGB-image guidance branch. On the one hand, considering the uncertainty and noise of coarse labels, a separate RGB-image guidance branch is introduced to form the two-stream structure and provide effective guidance information from the raw RGB data. On the other hand, we propose a BGA to achieve two-stage feature decoding, where the guidance stage aims to gain relatively robust baseline performance for mainstream branch by the guidance branch information, and the aggregation stage is to integrate the encoder features, previous decoder features, and global features by considering the roles of different features.
The S-Net is a replaceable salient object detection network supervised by the pseudo label\footnote{For clarity, the coarse label in our paper specifically refer to labels generated by unsupervised SOD model and used as input to R-Net. The pseudo label is saliency map obtained by testing with the trained R-Net or S-Net.} generated by the current R-Net, with the original RGB image as input and output of the pseudo label for the subsequent round learning of R-Net.
By using such a decoupled architecture, not only the negative impact of coarse labels on S-Net can be effectively reduced, but also the number of training samples can be expanded with the refined coarse labels generated by the trained R-Net, thereby enhancing the learning ability of the network.

Besides, in the face of imbalanced training on the hybrid-labeled data, some well-designed training strategies are crucial to guarantee the effectiveness and efficiency of network training.
Specifically, we design three ingenious training strategies:
(1) Alternate iteration mechanism. In order to ensure that sufficient and effective samples participate in training, we alternately perform iterative training of R-Net and S-Net. The two networks provide better labels for each other.
(2) Group-wise incremental mechanism. In order to avoid the imbalance of inputting a large number of pseudo-labeled samples and a small number of real-labeled samples at the same time, we group the training set and gradually increase the amount of data with pseudo labels in each training iteration, thereby stepwise learning the effective feature representations.
(3) Credibility verification mechanism. In order to ensure that the two networks can provide credible labels to each other during the iteration process, starting from the second iteration, we design a validation phase on the validation set containing 100 images, and only the best model that satisfies the validation conditions can be used to generate pseudo labels for the corresponding data to participate in the next step of training.
In general, the three training strategies constrain the training process from three aspects, \ie, quantity allocation, training method and reliability judgment, so as to achieve the effective training of the network.

The main contributions of this paper mainly lie in three aspects, including task setting, technical framework and training strategy:

\begin{itemize}
\item For the first time, we launch a new weakly-supervised SOD task based on hybrid labels, with a large number of coarse labels and a small number of real labels as supervision. To this end, we decouple this task into two sub-tasks of coarse label refinement and salient object detection, and design the corresponding R-Net and S-Net. Moreover, our method achieves competitive performance on five widely used benchmark datasets using only one-tenth of the real labels in fully-supervised setting.
\item 
We design a BGA in the R-Net to achieve two-stage feature decoding, where the guidance stage is used to introduce the guidance information from the RGB-image guidance branch to guarantee a relatively robust performance baseline, and the aggregation stage is to dynamically integrate different levels of features according to their modification or supplementation roles.

\item In order to guarantee the effectiveness and efficiency of network training, from the perspective of quantity allocation, training method and reliability judgment, we design the alternate iteration mechanism, group-wise incremental mechanism, and credibility verification mechanism.

\end{itemize}


\section{RELATED WORK}
\subsection{Fully supervised salient object detection}
Inspired by image semantic segmentation, Zhao \etal
~\cite{DBLP:conf/cvpr/ZhaoOLW15} proposed a fully supervised model based on CNN to integrate local and
global features to predict the saliency map. Wang \etal ~\cite{DBLP:conf/eccv/WangWLZR16}
adopted a recurrent CNN to refine the predicted saliency map step by step. Most of them follow an encoder-decoder architecture similar to FCN \cite{long2015fully}, on this basis, in order to obtain more accurate and convincing detection results, the researchers carried out a series of elaborate network designs.
Several recent works \cite{DBLP:conf/iccv/ZhangWLWR17,hou2017deeply,DBLP:conf/ijcai/DengHZXQHH18,DBLP:conf/aaai/HuZQFH18,DBLP:conf/cvpr/ZhangDLH018,DBLP:conf/cvpr/ZhangWQLW18,wei2020f3net,DBLP:conf/aaai/WangCZZ0G20,DBLP:conf/cvpr/PangZZL20,DBLP:conf/cvpr/WeiWWSH020,DBLP:conf/eccv/ZhaoPZLZ20} integrated features in multiple layers of CNN to exploit the context information at different semantic levels. Among them, Hou \etal ~\cite{hou2017deeply} introduced short connection to the skip-layer structure for capturing fine details. Deng \etal ~\cite{DBLP:conf/ijcai/DengHZXQHH18} proposed an iterative method to optimize the saliency map, leveraging features generated by deep and shallow layers. Zhang \etal ~\cite{DBLP:conf/cvpr/ZhangWQLW18} designed an attention guided network that selectively integrates multi-level contextual information in a progressive manner. Wei \etal ~\cite{wei2020f3net} focused on the feature fusion strategies and proposed a SOD network that equipped with cross feature module and cascaded feedback decoder trained with a new pixel position aware loss. Pang \etal ~\cite{DBLP:conf/cvpr/PangZZL20} investigated the multi-scale issue in salient object detection and proposed an effective and efficient network with the transformation-interaction-fusion strategy.
In the SOD task, for obtaining results with elaborate boundaries, edge-guided or boundary-guided methods have been proposed. Qin \etal ~\cite{feng2019attentive} proposed a boundary-aware model to segment salient object regions and predict the boundaries simultaneously. Li \etal ~\cite{zhao2019egnet} proposed an edge-guided SOD network to learn the complementarity between salient edge information and salient object information in a single network. Feng \etal ~\cite{feng2019attentive} designed an attentive feedback network by integrating some feedback network modules to explore the structure of objects better and proposed a new boundary-enhanced loss for learning exquisite boundaries. Wang \etal ~\cite{wang2019salient} introduced the salient edge detection module into an essential pyramid attention structure for salient object detection and achieved superior performance.

Although superior performance has been obtained, a fatal problem still exists is that they all require a mass of pixel-level labeled training data. Accordingly, how to obtain satisfactory detection results with fewer annotations has become a topic worth exploring, which also motivates our work.

\begin{figure*}[!t]
    \centering
    \includegraphics[scale=0.35]{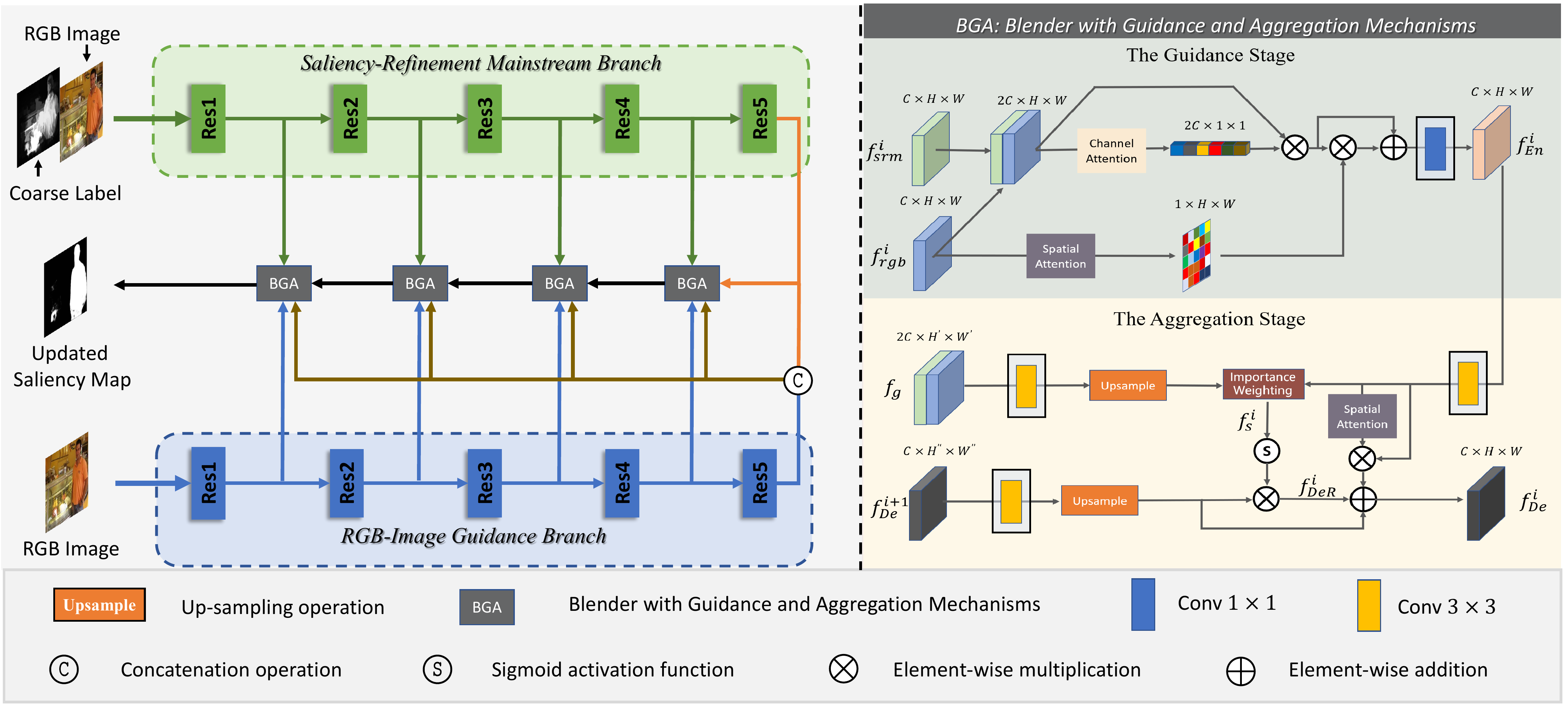}
    \caption{The overall framework of the proposed Refine Network (R-Net).}
    \label{fig:r-net}
\end{figure*}

\subsection{Weakly supervised salient object detection}
Unlike fully-supervised SOD needs a complete pixel-level label for each training sample, weakly-supervised SOD model may utilize simpler labels, \eg, scribble/point/image-level label, as supervision signals to achieve comparable performance. Due to the low cost of labels and considerable prospects, it has received more and more attention.
The WSS model \cite{wang2017learning} is the first weakly-supervised SOD method by using image-level label, which employs a global smooth pooling layer and a foreground inference scheme to make the network generate good prediction results even for unseen categories.
Zeng \etal ~\cite{zeng2019multi} utilized multiple labels (\ie, image-level labels and captions) to train a SOD model, and then a classification network and a caption generation network were designed to predict object class and generate captions, respectively.
Zhang \etal ~\cite{zhang2020weakly} used scribble labels as the supervision to train the network, including an auxiliary edge detection task to locate object edges explicitly and a gated structure-aware loss to place constraints on the scope of structure to be recovered.
Apart from that, Zhang \etal ~\cite{zhang2018deep} first applied an unsupervised method to generate coarse labels, then obtained refined saliency maps by modelling the noise in the coarse labels.
Zheng \etal ~\cite{DBLP:journals/tcsv/ZhengTZML21} first introduced saliency subitizing as the weak supervision and proposed a SOD model with saliency subitizing module and saliency updating module.
Yu \etal \cite{DBLP:conf/aaai/YuZXL21} proposed a local coherence loss
to propagate the labels to unlabeled regions based on image features and pixel distance. Piao \etal \cite{DBLP:conf/iccv/PiaoWZL21} introduced a new multiple-pseudo-label framework to integrate more comprehensive and accurate saliency cues from multiple labels, avoiding that the generated single label is inevitably affected by adopted refinement algorithms. Gao \etal \cite{DBLP:conf/aaai/Gao00GZHZ22} proposed a point supervised saliency detection model, where an adaptive masked flood filling algorithm is designed to generate pseudo labels, and the transformer-based point-supervised SOD model and a Non-Salient Suppression (NSS) method are used to achieve two-stage saliency map generation and optimization. Yan \etal\cite{DBLP:conf/aaai/YanWLZLL22} made the first attempt to achieve SOD by exploiting unsupervised domain adaption from synthetic data, and constructed a synthetic SOD dataset named UDASOD.

Besides, weakly supervised video object segmentation/SOD methods can also provide us with some enlightenment. Zhao \etal \cite{DBLP:conf/cvpr/Zhao0LBLH21} proposed the first weakly supervised video salient object detection model based on "fixation guided scribble annotations".
And some methods used weakly-supervised approaches to video object segmentation by fusing information between different frames \cite{DBLP:journals/tcsv/LinXLZ22,DBLP:journals/tist/WeiLLFWC22,DBLP:conf/aaai/LinX0021}. In contrast, Zhou \etal \cite{DBLP:conf/aaai/ZhouWZYL020} relied
only on the current frame image and the corresponding optical flow data to achieve the zero-shot video object segmentation.
En \etal \cite{DBLP:journals/tip/EnDZ21} performed video object segmentation with the help of saliency information.

In this paper, we construct a new label form for the first time to solve the weakly-supervised SOD task, namely hybrid labels, which only contains one-tenth of the real pixel-wise labelled samples. With the help of the proposed learning framework and training strategies, our method finally achieves encouraging performance.


\section{PROPOSED METHOD}
\subsection{Overview}
At first, the hybrid labels used in this paper can be divided into two parts, \ie, a small number of pixel-level real labels and a large number of coarse labels, where the coarse labels are generated by a traditional unsupervised method (\eg, MB \cite{zhang2015minimum}). The overall framework is shown in Fig. \ref{fig:general-framework}, consisting of a Refine Network (R-Net) and a Saliency Network (S-Net). The R-Net is designed as a two-stream encoder-decoder architecture that takes original RGB image and coarse label as inputs and outputs updated pseudo-labels (more details will be introduced in Section \ref{R-Net}). The S-Net is a replaceable SOD network that takes the original RGB image as input and the pseudo labels generated by R-Net as supervision signal.

The proposed framework is trained in alternating iterations. For the training process, the first thing to mention is that we divide the training samples into ten groups (note that only the samples in GROUP 1 include real labels) and incrementally load them into the training pool.
On the one hand, the alternating iteration strategy makes the quality of pseudo labels continuously be optimized through the cooperation of these two networks.
On the other hand, incremental loading of training samples enhances the guiding ability of real labels. As a result, the imbalance between the real label and pseudo label can be alleviated effectively.
Specifically, taking the first iteration as an example, we first use the original RGB images and coarse labels of GROUP 1 as inputs to train the R-Net, then predict the corresponding pseudo labels of GROUP 2. Next, we input the GROUP 1 and GROUP 2 into S-Net for network training, then predict the corresponding pseudo labels of GROUP 3.
At this point, this round of training is over.
The samples in GROUP 1 and GROUP 3 will be used for the next iteration of R-Net training.
The iteration is terminated until we run out of data.
Ultimately, we only use the trained S-Net for testing and no longer need coarse label input.
More details will be introduced in the Training Strategy with Hybrid Labels of Section \ref{training}.

\subsection{Refinement Network (R-Net)}\label{R-Net}
The R-Net is designed to refine the coarse labels and produce better pseudo labels that can be used for S-Net training. Intuitively, we only need to input coarse labels and corresponding RGB images into the network to achieve the label refinement. But in our setting, the coarse labels are generated by the traditional unsupervised method, which very noisy for some complex scenes, even inferior to the results obtained by the weakly-supervised deep learning methods. In this way, if only the Saliency-Refinement Mainstream Branch (mainstream branch for short) is used to directly refine the label, the difficulty can be imagined. Considering the uncertainty and noise of coarse labels, a separate RGB-Image Guidance Branch (referred to as the Guidance Branch) is introduced into the R-Net to form a two-stream encoding structure, which is used to provide some guidance information to the mainstream branch, such as object localization and integrity, thereby guaranteeing a relatively robust performance baseline.
The whole framework of the R-Net is shown in Fig. \ref{fig:r-net}.
The encoders of both streams are based on the ResNet-50 \cite{he2016deep} to extract the corresponding multi-level features. Then, we propose a Blender with Guidance and Aggregation Mechanisms (BGA) to achieve two-stage feature decoding, as shown on the right side of Fig. \ref{fig:r-net}.
The role of the first stage is guidance, that is, to supplement the mainstream branch with the information of the guidance branch, ensuring that the mainstream branch has a relatively robust baseline performance.
The role of the second stage is aggregation, that is, to integrate the encoder features, previous decoder features, and global features by considering the roles of different features.



\subsubsection{The Guidance Stage}
We hope that in the first stage, the RGB branch can provide guidance information (\eg, object localization and integrity) for the mainstream branch, guaranteeing its effective learning and robust performance baseline. The detailed architecture is shown on the top right corner of Fig. \ref{fig:r-net}.



First, in order to ensure that enough saliency information can be transferred to the mainstream branch and mitigate unreliable noise from the coarse label input, we supplement and filter the features in the channel dimension. Specifically, the encoder features of corresponding layers in the two branches are first concatenated for complementation, and then channel attention is used to highlight essential channel features for filtering. This process can be formulated as:
\begin{align}
    F_{com}^{i}=CA([f_{srm}^{i},f_{rgb}^{i}])\circledcirc[f_{srm}^{i},f_{rgb}^{i}],
\end{align}
where $F_{com}^{i}$ denote the complementation features after channel attention, $CA$ is the channel attention operation \cite{ca}, $f_{srm}^{i}$ and $f_{rgb}^{i}$ denote the encoder features of the ${{i}^{th}}$ layer in the mainstream branch and guidance branch, respectively,  $\left[ \cdot ,\cdot  \right]$ represents the concatenation operation along the channel dimension,
and $\circledcirc$ means element-wise multiplication with channel-wise broadcasting.

Secondly, in addition to the direct complement of channel dimensions, the RGB branch can also provide pixel-level spatial guidance information, which can both reinforce important regions and suppress irrelevant noise interference.
Specifically, we use the spatial attention \cite{sa} to generate the spatial location mask that need to be emphasized from the perspective of RGB information, and use this to update the features of the refinement branch.
\begin{align}
    F_{En}^{i}=Con{{v}_{1\times 1}}(SA(f_{rgb}^{i})\odot F_{com}^{i} + F_{com}^{i}),
\end{align}
where $F_{En}^i$ are the final output encoder features of the guidance stage after the spatial attention, $SA$ is the spatial attention operation \cite{DBLP:conf/eccv/WooPLK18},  $\odot$ is the element-wise multiplication, and $Con{{v}_{1\times 1}}$ denotes the convolutional layer with the kernel size of $1\times 1$.



\subsubsection{The Aggregation Stage}
As mentioned earlier, the second stage is mainly used to realize the fusion of multi-level features, including the encoder features of the corresponding layer generated in the first stage, the global features from the top encoder layer, and the decoder features of the previous layer.
In order to implement the aggregation more effectively, we need to analyze the roles of various features. In general, both encoder features and global features should play an auxiliary role to obtain better decoder features in the feature decoding stage.
The auxiliary functions can be divided into two aspects: one is to refine the decoder features under the guidance of global information; the other is to supplement the decoder features under the guidance of encoder features.

First, the high-level semantic features from the top encoder layer are crucial for distinguishing salient objects, but as the decoding process proceeds, the semantic constraint will be gradually diluted. Therefore, in order to enforce semantic information throughout the decoding process, we generate the corresponding semantic guidance mask to refine the decoder features of each level.
Specifically, we firstly combine the semantic features from two branches and the encoder features generated in the first stage through an importance weighting strategy \cite{CoADNet}:
\begin{align}
      f_{s}^i=P^i\odot f_{g}+(1-P^i)\odot f_{En}^{i},
\end{align}
where $f_{g}=conv([{f_{srm}^{5}},{f_{rgb}^{5}}])$ denote the fused semantic features from two branches, $P^i$ is the learned importance weight that controls the fusion rate of the features of $f_{g}$ and $f_{En}^{i}$ (more details can be found in \cite{CoADNet}). Then, the fusion features $f_{s}^i$ containing global semantic information are activated as a semantic mask, which is used to modify the upsampled decoder features:
\begin{align}
  f_{DeR}^{i}=Up(f_{De}^{i+1})\odot \sigma (f_{s}^{i}),
\end{align}
where $f_{DeR}^{i}$ are the modified decoder features of the $i^{th}$ level, $f_{De}^{i+1}$ represent the original decoder features of the $(i+1)^{th}$ level, $Up$ represents the up-sampling operation by bilinear interpolation, and $\sigma$ is the sigmoid activation function.

Second, as demonstrated in \cite{hou2017deeply}, the encoder features contain many valuable information that can complement the decoder feature learning, such as the shallower features including rich spatial information to better recover details, \etc.
Therefore, we further supplement the modified decoder features with the filtered features via the spatial attention mechanism \cite{sa} to obtain more comprehensive saliency-related decoder features.
This process can be formulated as:
\begin{align}
  f_{De}^{i}=Up(f_{De}^{i+1})+f_{DeR}^{i}+SA(f_{En}^{i})\odot f_{En}^{i},
\end{align}
where $f_{DeR}^{i}$ are the modified decoder features of the $i^{th}$ level, and $SA$ is the spatial attention operation \cite{sa}.

\subsection{Training Strategy with Hybrid Labels}\label{training}

\begin{figure}[!t]
    \centering
    \includegraphics[scale=0.45]{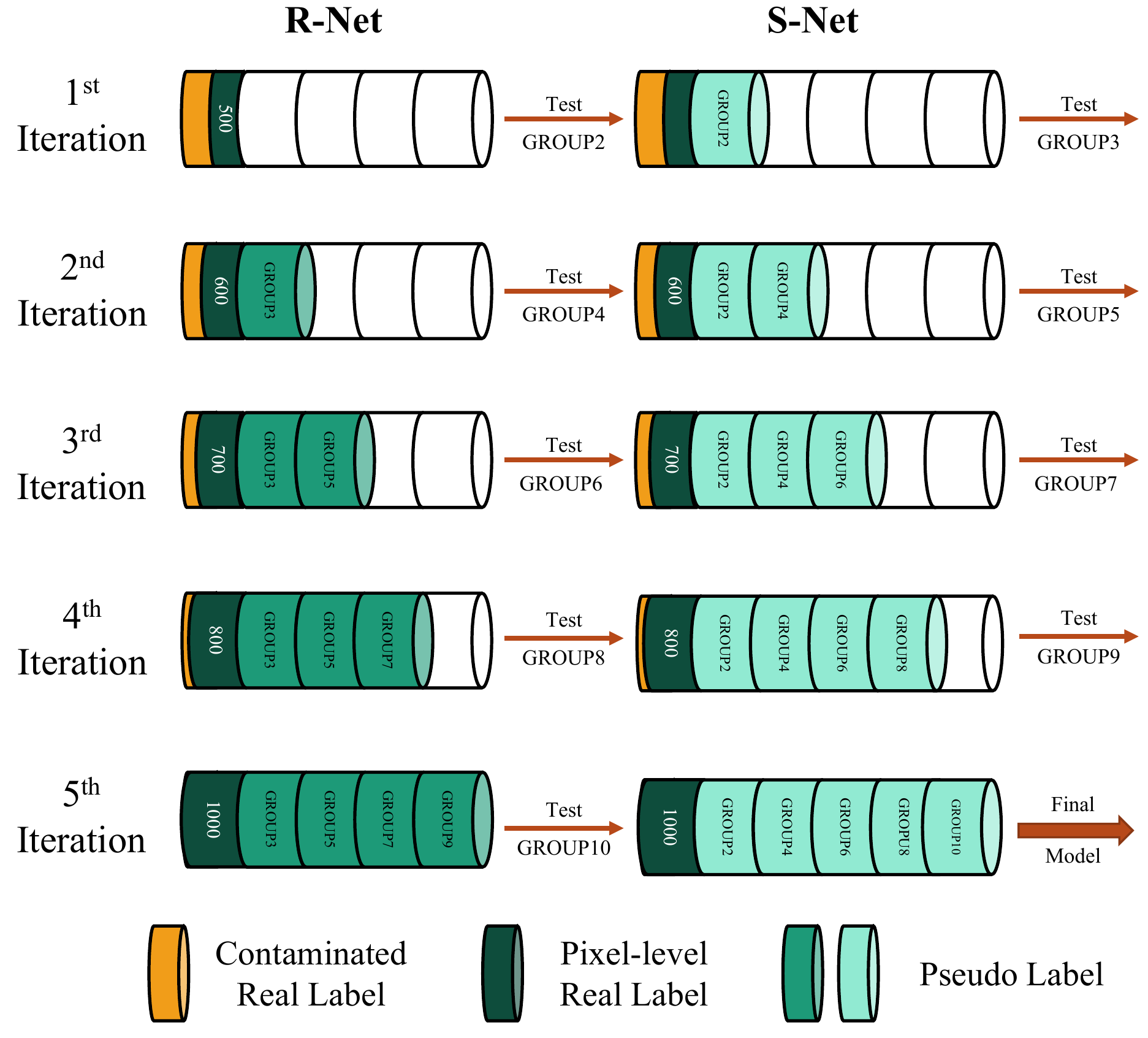}
    \caption{Training strategy for group update based on real labels and pseudo labels.}
    \label{fig:my_label}
\end{figure}

\textbf{Training settings.} The pixel-level real labels and coarse labels are given in our training set, in which the coarse labels are only used as the input to the S-Net instead of supervision.
At the same time, the pseudo label will be generated as supervision information during the network training process.
In the implementation, we randomly select 1,000 samples from the DUTS-TR dataset \cite{wang2017learning} as the real-labeled training subset, and use the MB method \cite{zhang2015minimum} to generate the corresponding coarse labels of all samples in the DUTS-TR dataset (including the 1,000 samples mentioned earlier).
As thus, these 1,000 samples (including RGB images, coarse labels, and real labels) can support the first iteration of R-Net training.
In order to guarantee the effectiveness and efficiency of network training, we propose three key training mechanisms, including alternate iteration mechanism, group-wise incremental mechanism, and credibility verification mechanism, as illustrated in Fig. \ref{fig:my_label}.

\textbf{Alternate iteration mechanism.}
As mentioned earlier, considering that coarse labels may contain a lot of noise, directly training the network under these supervisions will inevitably lead to poor performance.
Therefore, we design a R-Net for label correction and a S-Net for salient object detection.
In terms of network training, we train these two networks alternately and iteratively, thereby providing better pseudo labels for each other.
In detail, the S-Net of the current iteration is trained using the pseudo labels generated by the trained R-Net of the current iteration, and the pseudo labels generated by the trained S-Net are further used for the next iteration of R-Net training.
The two networks are trained in an alternating manner until all training samples are traversed, which is called alternate iteration mechanism.


\textbf{Group-wise incremental mechanism.}
Another important problem in the weakly-supervised SOD framework with hybrid label is sample imbalance caused by the difference in the number of real-labeled samples and coarse-labeled samples.
If the unbalanced training samples are directly used for network training, it will cause ambiguity and unavailability of network learning. Therefore, we propose a group-wise incremental mechanism to avoid network collapse caused by importing a large amount of pseudo-labeled data at a time. Specifically, we divide all training samples (\ie, 1,000 real-labeled samples and 9,000 coarse-labeled samples) into ten groups equally, of which 1,000 samples with real labels are grouped into GROUP 1.
In the first iteration, we only load the samples of GROUP 1 to train the R-Net, and then we use the trained R-Net to test the samples in GROUP 2 and obtain the the corresponding pseudo labels.
Subsequently, all the samples in GROUP 1 with real labels and GROUP 2 with pseudo labels are used for S-Net training.
Finally, the trained S-Net is used to test the samples in GROUP 3, and the generated saliency maps are used as the pseudo labels for the next iteration of R-Net training. At this point, the first iteration is completed.
Noteworthy, in order to prevent the model from overfitting to real-labeled data and improve the robustness of the model, we do not load all real-labeled samples into training in the first iteration, but also use an incremental strategy.
To be specific, in the first iteration, we select 500 samples from GROUP 1 as the real-labeled sample batch, and the remaining 500 samples are degenerated into contaminated-labeled sample batch through the rotation, cropping, and occlusion operations. In subsequent training iterations, we gradually reduce the number of contaminated-labeled data and increase the amount of real-labeled data.
In the second iteration, we still train R-Net first, followed by the S-Net.
For the R-Net training, 2,000 samples in GROUP 1 and GROUP 3 are used, where the number of real-labeled samples is increased to 600 and the number of contaminated-labeled samples is decreased to 400.
Note that, the corresponding pseudo labels in GROUP 3 are obtained by testing the samples in GROUP 3 using the S-Net trained in the previous iteration.
Then, the new trained R-Net is utilized to test the GROUP 4. And the samples in GROUPs 1, 2, 4 are used to train the S-Net again.
In order to traverse the entire training dataset (\ie, 10,000 samples), 5 iterations need to be performed, and the training process for other iterations can be analogized.
In addition, since both R-Net and S-Net are trained with hybrid labels, in order to ensure the effectiveness and performability of training, we first train the network under the pseudo-labeled samples and then fine-tune the real-labeled samples in a training epoch.

\textbf{Credibility verification mechanism.}
The purpose of alternating training of the two networks is to provide better pseudo labels for each other, so we introduce a credibility verification mechanism from the second iteration to ensure the validity of the provided labels.
Only when the current model outperforms the previous best model on the validation set including 100 images, we use it to generate pseudo-labels for the corresponding group and participate in the next training step.
Taking the validation process of S-Net as an example, in the second iteration, if the MAE score of the newly trained R-Net model using GROUP1 and GROUP3 data on the validation set is smaller than the MAE score \footnote{MAE refers to the mean absolute error, which represents the error between the prediction and the ground truth, and the smaller the value means the better.} of the previous best R-Net model (the R-Net model trained in the first iteration at this time), then we use the current R-Net model to test the GROUP 4 and generate the corresponding pseudo labels, otherwise we use the R-Net model trained in the first iteration to generate the pseudo labels.
Validation of other iterations is similar to this process.

 \begin{figure*}[!t]
	\centering
	\includegraphics[scale=0.68]{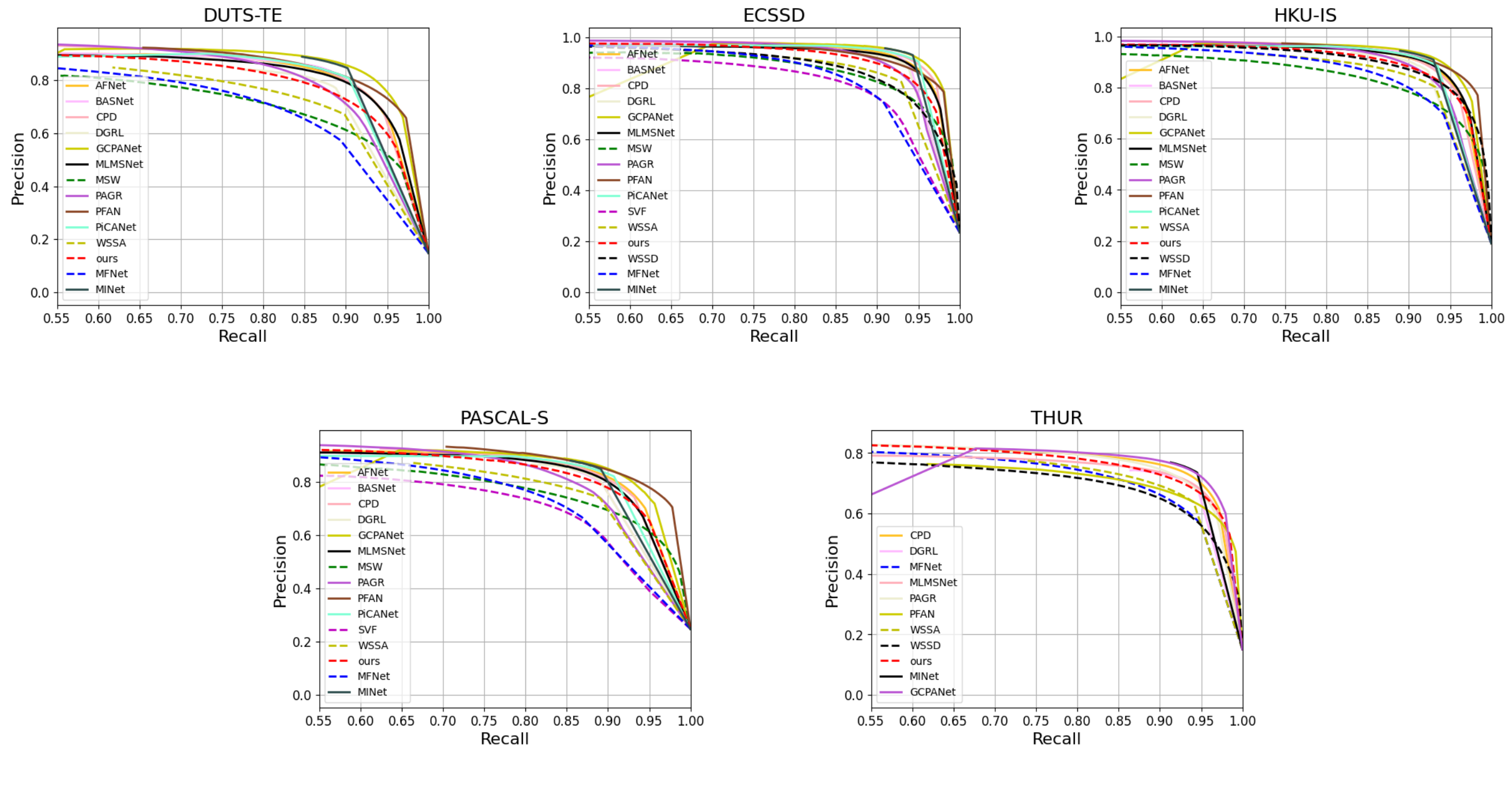}
	\caption{PR curves curves  on five common saliency datasets. Solid lines are fully-supervised methods, dashed lines are weakly-supervised and unsupervised  methods.}
	\label{pr_fm}
\end{figure*}

\subsection{Loss Function}

Referring to the traditional SOD method, we also use the binary cross-entropy loss as the loss function for R-Net training.
As mentioned earlier, in order to reduce noise pollution from pseudo labels and enhance the guidance of real labels, we first train the network under the pseudo-labeled samples and then fine-tune the real-labeled samples in a training epoch.
Specifically, we treat these two types of labels differently and change the binary cross-entropy loss to the following form:
\begin{align}
    \footnotesize
    {\mathcal{l}_{r}}=-\sum\limits_{j\in {{D}^{r}}}{[Y_{j}^{r}\log R(X_{j}|\Phi )-(1-Y_{j}^{r})\log (1-R(X_{j}}|\Phi ))],
\end{align}
\begin{align}
    \footnotesize
    {\mathcal{l}_{p}}=-\sum\limits_{k\in {{D}^{p}}}{[Y_{k}^{p}\log R(X_{k}|\Phi )-(1-Y_{k}^{p})\log (1-R(X_{k}}|\Phi ))],
\end{align}
where ${\mathcal{l}_{r}}$ and $\mathcal{l}_{p}$ are all standard BCE losses, but the samples used in the calculation of the two loss functions are different. The loss ${\mathcal{l}_{r}}$ calculates the BCE loss of samples with real labels, while $\mathcal{l}_{p}$ calculates the BCE loss of samples with pseudo labels.
 $D^r$ and $D^p$ correspond to the training set with real labels and pseudo labels, respectively. $\{X,Y\}$ denotes the training sample in the corresponding set, where $X$ are the inputs of the R-Net including the RGB image and the corresponding coarse label, and $Y$ is the real label or pseudo label of the sample.
$R(\cdot|\Phi )$ denotes the R-Net, and $\Phi$ represents the network parameters of R-Net.



\begin{table*}[!t]
\setstretch{1}
\renewcommand{\arraystretch}{1.4}
\caption{Quantitative results of different methods on five SOD benchmark datasets, $\uparrow$ and $\downarrow$ respectively indicate that the larger and smaller the score, the better. `F' means fully supervision, `I’ means image-level weakly supervision, and `S' means scribble-level weakly supervision, `Sub' means subitizing supervision, `M’ means multi-source weakly supervision, `Un’ is for unsupervision, and `H’ denotes hybird supervision. The best performance is marked in \textbf{BOLD}, and the second best performance is marked in \underline{UNDERLINE}.}
\resizebox{\textwidth}{45mm}{
\begin{tabular}{ccc|ccc|ccc|ccc|ccc|ccc}\hline
        &      &     & \multicolumn{3}{c|}{DUTS-TE} & \multicolumn{3}{c|}{ECSSD}   & \multicolumn{3}{c|}{HKU-IS}  & \multicolumn{3}{c|}{PASCAL-S} & \multicolumn{3}{c}{THUR} \\ \hline
        & SUP & YEAR & ${{F}_{\beta }^{max}}\uparrow $ & ${{S}_{m}}\uparrow $ & $MAE\downarrow $   & ${{F}_{\beta }^{max}}\uparrow $ & ${{S}_{m}}\uparrow $ & $MAE\downarrow $   & ${{F}_{\beta }^{max}}\uparrow $ & ${{S}_{m}}\uparrow $ & $MAE\downarrow $   & ${{F}_{\beta }^{max}}\uparrow $  & ${{S}_{m}}\uparrow $ & $MAE\downarrow $   & ${{F}_{\beta }^{max}}\uparrow $  & ${{S}_{m}}\uparrow $  & $MAE\downarrow $   \\ \hline
DGRL    & F  & 2018       & 0.805    & 0.842    & 0.050  & 0.913    & 0.903    & 0.041 & 0.900     & 0.894    & 0.036 & 0.837      & 0.836    & 0.072 & 0.746     & 0.813     & 0.076 \\
PiCANet & F  & 2018       & 0.840    & 0.863    & 0.040  & 0.928    & 0.916    & \textbf{0.035} & 0.913    & 0.905    & \underline{0.031} & 0.848     & 0.846    & \underline{0.065} & -     & -     & - \\
PAGR    & F  & 2018       & 0.816    & 0.838    & 0.056 & 0.904    & 0.889    & 0.061 & 0.897    & 0.887    & 0.048 & 0.822     & 0.819    & 0.092 & 0.769     & 0.830     & 0.070 \\
MLMSNet & F  & 2019       & 0.825    & 0.861    & 0.049 & 0.917    & 0.911    & 0.045  & 0.910    & 0.906    & 0.039 & 0.841     & 0.845     & 0.074 & 0.752     & 0.819     & 0.079 \\
CPD     & F  & 2019      & 0.840    & 0.869    & 0.043 & 0.926   & 0.918    & 0.037 & 0.911    & 0.905    & 0.034 & 0.842     & 0.847    & 0.072 & 0.774     & 0.834     & \underline{0.068} \\
AFNet   & F  & 2019       & 0.836    & 0.867    & 0.046 & 0.924    & 0.913    & 0.042 & 0.909    & 0.905    & 0.036 & 0.848     & 0.849    & 0.071  & -     & -     & - \\
BASNet  & F  & 2019       & 0.838    & 0.866    & 0.048 & 0.931     & 0.916    & 0.037 & 0.919    & 0.909    & 0.032 & 0.842     & 0.836    & 0.077 & -     & -    & - \\
PFAN    & F  & 2019       & 0.850    & 0.874    & 0.041 & 0.914    & 0.904    & 0.045 & 0.918    & \underline{0.914}    & 0.032 & \textbf{0.866}     & \underline{0.862}    & \underline{0.065} & 0.722     &  0.781     &  0.104 \\
GCPANet & F  & 2020       & \textbf{0.866}    & \textbf{0.891}    & \textbf{0.038} & \underline{0.936}    & \textbf{0.927}    &  \textbf{0.035} & \textbf{0.926}    &  \textbf{0.920}     &  \underline{0.031} & \underline{0.859}     &  \textbf{0.866}    &  \textbf{0.062} & \textbf{0.784}     & \textbf{0.840}     & 0.070 \\
MINet   & F  & 2020       & \underline{0.863}    & \underline{0.881}    & \underline{0.039}  &  \textbf{0.937}    & \underline{0.923}    & \underline{0.036} &  \underline{0.922}    & \underline{0.914}    & \textbf{0.030} &  0.856     & 0.855    & \textbf{0.062} &  \underline{0.778}      & \underline{0.836}     & \textbf{0.066}  \\ \hline
SVF     & Un  & 2017      & -        & -        & -     & 0.832    & 0.832    & 0.091 & -        & -        & -     & 0.734       & 0.757    & 0.134 & -     & -     & - \\
MNL     & Un  & 2018      & 0.725    & -        & 0.075 & 0.810     & -        & 0.091 & 0.820     & -        & 0.065 & 0.747     & -        & 0.157 & -     & -         & - \\
WSS     & I  & 2017       & 0.633    & -        & 0.100   & 0.767    & -        & 0.108 & 0.773    & -        & 0.078 & 0.697     & -        & 0.184 & -     & -         & -  \\
ASMO    & I   & 2018      & 0.568    & -        & 0.115 & 0.762    & -        & 0.068 & 0.762    & -        & 0.088 & 0.653     & -        & 0.205 & -     & -         & - \\
MSW     & M   & 2019      & 0.705    & 0.752    & 0.091 & 0.851     & 0.820    & 0.099 & 0.828    & 0.812    & 0.086 & 0.759     & 0.762    & 0.136 & -     & -     & - \\
MFNet     & I   & 2021      & 0.733    & 0.775    & 0.076 & 0.858     & 0.835    & 0.084 & 0.859    & 0.847    & 0.058 & 0.764     &0.768    & 0.117 & 0.731     & 0.795     & \underline{0.075} \\
WSSD     & Sub   & 2021      & -    & -    & - &\underline{0.873}     & 0.827    & 0.119 & \underline{0.884}    & \underline{0.870}    & 0.082 & \underline{0.820}    &\underline{0.814}    & 0.128 & 0.703     & 0.768     & 0.114 \\
WSSA    & S   & 2020      & \underline{0.755}    & \underline{0.803}    & \underline{0.062} &  0.871     & \underline{0.865}    &  \underline{0.059} & 0.864     & 0.865    & \underline{0.047} & 0.788     & 0.796    & \underline{0.094} &  \underline{0.736}     &  \underline{0.800}      &  0.077 \\
Ours    & H  &   &  \textbf{0.803}    &  \textbf{0.837}    &  \textbf{0.050} & \textbf{0.899}    &  \textbf{0.886}    & \textbf{0.051} &  \textbf{0.892}    &  \textbf{0.887}    &  \textbf{0.038} &  \textbf{0.827}     &  \textbf{0.828}    &  \textbf{0.076} & \textbf{0.755}      & \textbf{0.813}     & \textbf{0.069} \\ \hline
\end{tabular}}
\label{table-compare_value}
\end{table*}

In general, the whole loss of the R-Net consists of the dominant loss ${\mathcal{l}}_{dom}$ on the final prediction and three auxiliary losses $\mathcal{l}_{aux}^i$ on the side outputs generated by the middle three layers of the decoder, which is formulated as:
\begin{equation}
    {{L}_{R}}={{\mathcal{l}}_{k,dom}}+\sum\limits_{i=1}^{3}{{{\lambda }_{i}}\mathcal{l}_{k,aux}^{i}},
\end{equation}
where $k=\{r,p\}$ indexes the real-labeled data or pseudo-labeled data, and ${\lambda}_{i}$ are the hyper-parameters that control the weight of each auxiliary loss, which are set to $(0.2, 0.4, 0.8)$ in experiments.

Since the S-Net in this paper is replaceable, we still follow the loss function of the original paper in the training order of `first real labels, then pseudo labels'.

\section{EXPERIMENT}
\subsection{Implementation Details and Setup}
\subsubsection{Datasets}
Five widely-used salient object detection benchmark datasets are employed to evaluate the entire performance, including:
\begin{itemize}
    \item DUTS \cite{wang2017learning} dataset contains 10,553 training images (DUTS-TR) and 5,019 testing images (DUTS-TE), with pixel-wise saliency ground truth.
    \item ECSSD \cite{yan2013hierarchical} dataset consists of real images of complex scenes, containing 1,000 complex images with the corresponding the pixel-wise saliency ground truths.
    \item HKU-IS \cite{li2015visual} dataset includes 4,447 challenging images, most of which are low-contrast or have multiple salient objects.
    \item PASCAL-S \cite{li2014secrets} dataset consists of 850 images from the PASCAL VOC 2010 validation set, with multiple salient objects in the scene.
    \item THUR \cite{DBLP:journals/vc/ChengMHH14} dataset collects 15,000 images from the Internet and annotates each image with the corresponding pixel-level saliency ground truth.

\end{itemize}

\subsubsection{Evaluation Metrics}
We adopt Precision-Recall (PR) curve  \cite{crm2019tip,crm2020going},  max F-measure score \cite{crm2018tip,crm2019tc}, S-measure score \cite{fan2017structure}, and MAE score \cite{crmICME,crm2019tmm} as the evaluation metrics.
As the PR curve is closer to the upper right corner, the model performance is better. The larger the F-measure and S-measure values, the better the performance, while the MAE score is just the opposite.

\subsubsection{Implementation Details}
We select the first 1,000 samples in the DUTS-TR dataset \cite{wang2017learning} as the real-labeled data, providing the pixel-wise real ground truth.  Then, we use the MB method \cite{zhang2015minimum} to generate the saliency maps for all the images in the DUTS-TR dataset \cite{wang2017learning}, thereby forming the coarse-labeled set. The validation set includes a total of 100 images from the SOD dataset\cite{DBLP:conf/cvpr/MovahediE10}.

We use the Pytorch toolbox to implement the proposed network and accelerate training by an NVIDIA GeForce RTX 3090 GPU card. We also implement our network by using the MindSpore Lite tool\footnote{\url{https://www.mindspore.cn/}}.
The ResNet-50 is used as the backbone of the R-Net to extract encoding features, with initial parameters loaded from the pre-trained model on ImageNet \cite{deng2009imagenet}. The MINet \cite{DBLP:conf/cvpr/PangZZL20} is used as S-Net in our implementation.
For the R-Net, the training images are first resized to $288\times 288$ by uniform resizing and random cropping. All training samples are then augmented using random horizontal flips and rotations.
For the S-Net, we directly resize all images to $320\times 320$ during training and inference, and then apply the same augmentation strategy to R-Net.
During training, the R-Net and S-Net are optimized by Adam optimizer with the batch size of 8, momentum of 0.9, and the weight decay of $5e^{-4}$. The initial learning rate is set to $1e^{-4}$, and divided by ten every ten epochs.
The overall network is trained for a total of five iterations, and we design the same number of epochs (\ie, 30 epochs) for each iteration of the training process, whether it is R-Net or S-Net.
Note that we only use the warm-up strategy in the first iteration. In the R-Net, we need to concatenate the RGB image and coarse label together into the backbone, for a total of 4 channels. Following the setting in \cite{DBLP:conf/cvpr/FuFJZ20}, we duplicate the original three channels of the ResNet model and its parameter once to form 6 channels, and then take the first four channels as the input layer of the new model.

 \begin{figure*}[!t]
	\centering
	\includegraphics[scale=0.82]{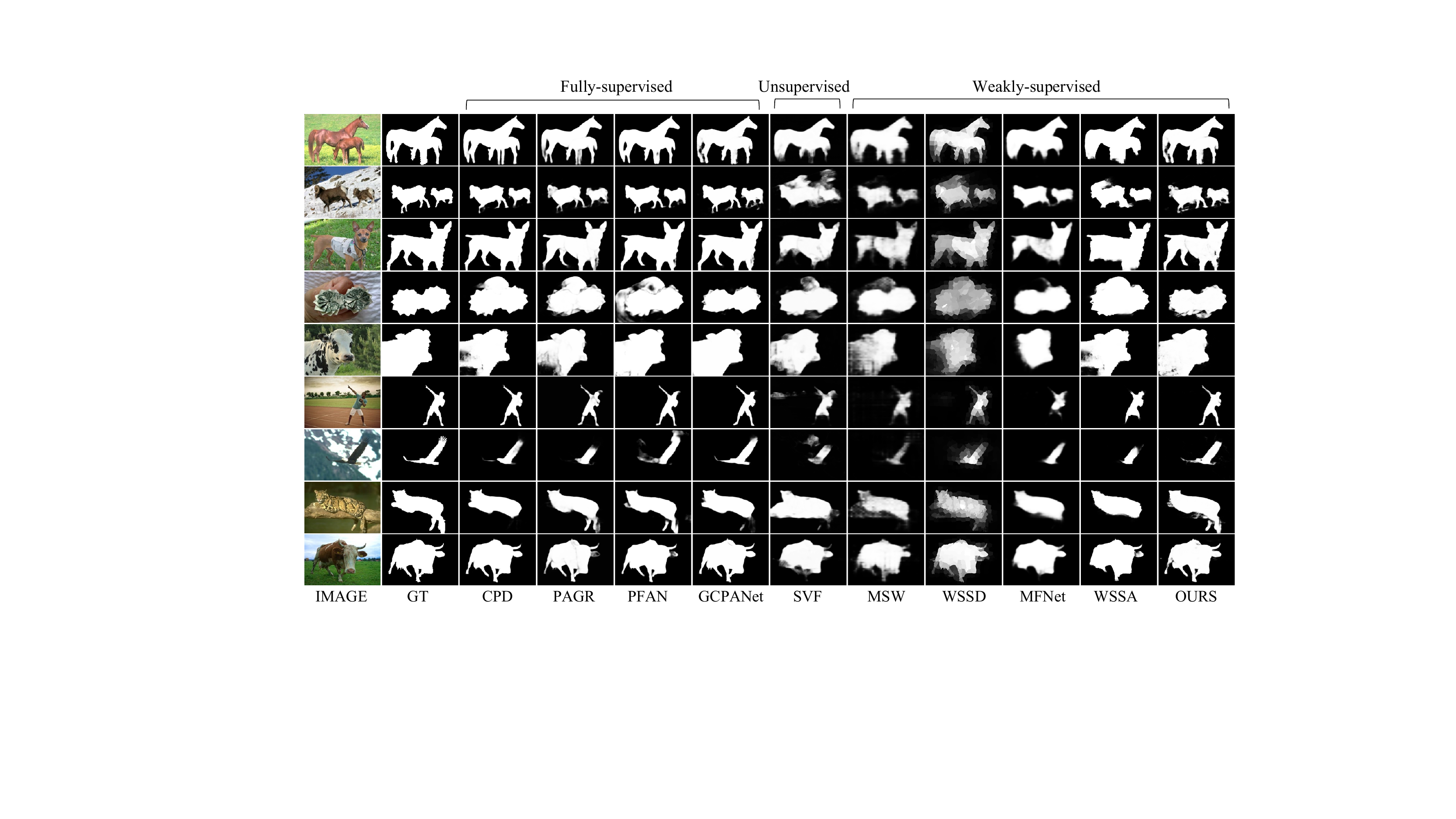}
	\caption{Visual comparisons with other state-of-the-art methods in various representative scenes.}
	\label{visualization_compare}
\end{figure*}

\subsection{Comparison with State-of-the-arts}
We compare the proposed method with other state-of-the-art models, including fully-supervised methods (\ie, PAGR \cite{zhang2018progressive}, MLMSNet \cite{wu2019mutual}, CPD \cite{wu2019cascaded}, AFNet \cite{feng2019attentive}, BASNet \cite{qin2019basnet}, GCPANet \cite{chen2020global}, DGRL \cite{DBLP:conf/cvpr/WangZWL0RB18}, PiCANet \cite{DBLP:journals/tip/LiuHY20}, PFAN \cite{DBLP:conf/cvpr/ZhaoW19}, and MINet \cite{DBLP:conf/cvpr/PangZZL20}), weakly-supervised methods (\ie, MSW \cite{zeng2019multi}, WSSA \cite{zhang2020weakly}, ASMO \cite{li2018weakly},  WSS \cite{wang2017learning}, MFNet \cite{DBLP:conf/iccv/PiaoWZL21}, and WSSD \cite{DBLP:journals/tcsv/ZhengTZML21}) and unsupervised methods (\ie, SVF \cite{zhang2017supervision} and MNL\cite{zhang2018deep}). For a fair comparison, the saliency maps of the different methods are provided by the authors or obtained by running the released code with default parameters.


\subsubsection{Quantitative Evaluation}
First of all, the PR curves are shown in Fig. \ref{pr_fm}. Our method (red dashed line) achieves the best performance in most cases compared to other weakly supervised and unsupervised methods on five common datasets, which is consistent with the quantitative scores reported in Table \ref{table-compare_value}.
Compared with the unsupervised SVF method \cite{zhang2017supervision} on the PASCAL-S dataset, the percentage gain of our method reaches $12.7\%$ for max F-measure, $9.4\%$ for S-measure and $43.3\%$ for MAE score.
Compared with the WSSD method with subitizing supervision \cite{DBLP:journals/tcsv/ZhengTZML21} and MFNet method with image-level supervision \cite{DBLP:conf/iccv/PiaoWZL21} on the HKU-IS dataset, the percentage gain reaches $53.7\%$ and $34.5\%$ for MAE score.

In addition, our method also achieves more competitive performance against the weakly-supervised SOD models with stronger supervision.
For example, compared with the MSW method \cite{zeng2019multi} with a variety of combination supervision labels, the percentage gain of S-measure reaches $11.3\%$ on the DUTS-TE dataset, and the the percentage gain of max F-measure also wins $13.9\%$.
For the scribble based weakly-supervised SOD method (\eg, WSSA \cite{zhang2020weakly}), although the proportion of annotations is relatively small, each sample clearly defines the foreground and background regions. That is, the supervision information given by the scribble is perfectly accurate.
By contrast, the hybrid labels we use contain $90\%$ coarse labels with a lot of uncertain noise, but our model outperforms the WSSA method overall on all metrics across all datasets.
For example, compared with the WSSA \cite{zhang2020weakly} on the DUTS-TE dataset, the percentage gains of max F-measure, S-measure, and MAE score reach $6.4\%$, $4.2\%$, and $7.1\%$, respectively.
It is worth mentioning that our method catches up or even surpasses some fully-supervised methods on certain datasets (\eg, THUR dataset).
In our proposed framework, we choose the MINet \cite{DBLP:conf/cvpr/PangZZL20} as our S-Net for training, and achieve the original performance of $80\%\sim90\%$ using only $1/10$ of the original training set. Of course, there is still a lot of room for improvement in the performance.

\subsubsection{Qualitative Comparison} Some visual comparisons are shown in Fig. \ref{visualization_compare}. It can be seen that our method surpasses the unsupervised and weakly-supervised methods in terms of structural integrity and accuracy, and achieves comparable results to fully supervised methods. Our advantages are reflected in the following aspects:
\begin{itemize}
    \item \emph{Advantages in background suppression}: Our model can effectively suppress noise and accurately locate the position of salient objects. For example, in the second image, the unsupervised SVF method \cite{zhang2017supervision} and scribble-based WSSA method \cite{zhang2020weakly} fail to accurately locate the boundary between the snow and the sheep from the complex backgrounds. Also, in the fourth image,
   some weakly-supervised methods (\eg, WSSD \cite{DBLP:journals/tcsv/ZhengTZML21}, WSSA \cite{zhang2020weakly}), as well as several fully-supervised methods (\eg, CPD \cite{wu2019cascaded}, PAGR \cite{zhang2018progressive}, PFAN \cite{DBLP:conf/cvpr/ZhaoW19}), are wrongly detect the hand as the salient object. In contrast, our model has better results in terms of location accuracy and background suppression.

    \item \emph{Advantages in detail depiction}: Our model has a better ability to capture detailed information such as sharp boundaries and complete structure.
    In the third image, other weakly-supervised methods either fail to detect the dog's limbs completely, or fail to distinguish the boundary between the limbs and the grass background.
    Similarly, neither the cow in the fifth image nor the person in the sixth image can be detected completely.
    In contrast, our method is not only able to detect the relatively complete structure of these salient objects, but also has clearer and sharper boundaries.
    In the last image, our method has a clear advantage in characterizing the horns and limbs of the cow compared to other weakly-supervised methods, especially our method can accurately detect the cow feet at a distance, producing more complete result.

    \item \emph{Advantages in low-contrast scene}: Our model can identify the salient object although in low-contrast scenes. For example,
    The color of the eagle¡¯s wings and the mountain behind it are so close that it is difficult for even fully-supervised methods to fully detect this region, such as CPD \cite{wu2019cascaded} and PAGR \cite{zhang2018progressive}.
    As you can easily imagine, all other weakly-supervised methods also fail in this region.
    Fortunately, thanks to the entire network architecture and the multi-dimensional feature fusion, the method proposed in this paper successfully detects the left wing of the eagle.
    Furthermore, in the eighth image, not only is the color of the leopard very close to the tree trunk, but part of its legs are covered by the tree trunk, which increases the difficulty of detection. However, our model can still detect the entire leopard's legs based on the relationship between objects, which even exceeds the detection ability of some strongly supervised models.

\end{itemize}


\subsection{Ablation Study}
To validate the effectiveness of our proposed network, we conduct comprehensive ablation experiments on the HKU-IS, DUTS-TE, and PASCAL-S datasets, including the overall framework, the design of R-Net, and the training strategy.

\begin{table}[!t]
\centering
\renewcommand\arraystretch{1.2}
    \begin{spacing}{1}
    \caption{The effectiveness analyses of overall framework on the PASCAL-S, DUTS-TE and HKU-IS datasets.}
    \label{as-1}
    \setlength{\tabcolsep}{0.6mm}{
        \begin{tabular}{c|cc|cc|cc}
        \hline
                      & \multicolumn{2}{c|}{PASCAL-S} & \multicolumn{2}{c|}{DUTS-TE}
                      & \multicolumn{2}{c}{HKU-IS}\\ \cline{2-7}
                     & ${{F}_{\beta }^{max}}\uparrow$      & $MAE\downarrow $
                     & ${{F}_{\beta }^{max}}\uparrow$      & $MAE\downarrow $
                     & ${{F}_{\beta }^{max}}\uparrow$      & $MAE\downarrow $  \\ \hline
        M1           & 0.690            &0.161             & 0.622                & 0.136               & 0.775              &  0.101           \\
        M2           & 0.783            & 0.120            &  0.741               & 0.089               & 0.855              &  0.070           \\
        M3& 0.801 & 0.093 &  0.755    & 0.065    &  0.868   & 0.048           \\
        Ours        &0.827            & 0.076           & 0.803     &  0.050   &0.892 & 0.038             \\ \hline
        \end{tabular}}
    \end{spacing}
\end{table}
\renewcommand\arraystretch{2}
\begin{figure}[!t]
	\centering
	\includegraphics[scale=0.315]{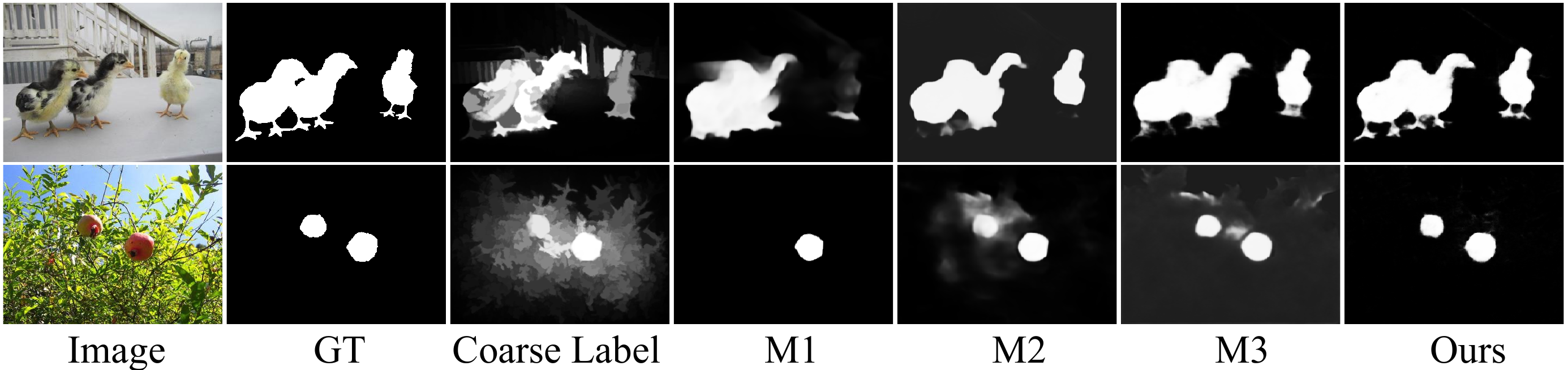}
	\caption{Visualization results for effectiveness of the overall framework.}
	\label{visualization-as-1}
\end{figure}
\subsubsection{Effectiveness of the overall framework}
In the face of hybrid labels, Fig. \ref{fig:general-framework} shows two SOD framework pipelines. One is direct hybrid training, as shown in Fig. \ref{fig:general-framework}(a), or even training with only real labels, and the other is our proposed framework in this paper, where label refinement and SOD are alternately trained under hybrid labels. To verify the effectiveness of our overall framework, we design three ablation experiments.
(1) M1: we first train the S-Net with 9,000 coarse-labeled samples and then fine-tune it with 1,000 real-labeled samples.
(2) M2: we only train the S-Net with 1,000 real-labeled samples.
(3) M3: we first train the S-Net using 1,000 samples with real labels, then use the trained S-Net to predict and update the original coarse labels for the remaining 9,000 samples, and finally retrain the S-Net by using the updated coarse-labeled samples and real-labeled samples.
The quantitative results on the PASCAL-S, DUTS-TE and HKU-IS datasets are reported in Table \ref{as-1}, and some visual comparisons are shown in Fig. \ref{visualization-as-1}.

Comparing M1 and M2, we can see that directly introducing coarse labels leads to a significant drop in performance, mainly due to the unreliable noise of coarse labels.
Compared with these two schemes, our proposed framework guarantees that better results can still be achieved when training with coarse labels.
For example, on the DUTS-TE dataset, the percentage gain of max F-measure against the M2 model is 8.4\%, and the percentage gain of max F-measure against the M1 model reaches 29.1\%. All these results demonstrate the effectiveness of our overall framework. As can be seen from Fig. \ref{visualization-as-1}, experiment M1 can only detect the main part of the object, with obvious omissions (such as the chicken on the far right in the first image and the pomegranate on the left in the second image). Furthermore, the results of experiment M1 are inferior to those of experiment M2 trained with only 1,000 real-labeled data. However, the M2 still contains a lot of noise and has very limited ability to describe the details (such as the duck's paws in the first image).

In addition, the performance of experiment M3 outperforms the experiment M2, but our model still has obvious advantages in performance. For example, on the DUTS-TE dataset, compared with experiment M3, the max F-measure score of the full model is improved from 0.755 to 0.803 with a percentage gain of 6.3\%, and the MAE score is improved from 0.065 to 0.050 with a percentage gain of 23.1\%. On the HKU-IS dataset, the max F-measure score of the full model is improved from 0.868 to 0.892 with a percentage gain of 2.8\% compared with experiment M3, and the MAE score is improved from 0.048 to 0.038 with a percentage gain of 20.8\%.
For the experiment M3, although the noise is significantly reduced, the performance is still inferior to our framework in terms of details, such as chicken feet.
All these experiments verify the effectiveness of our overall framework.



\renewcommand\arraystretch{1.4}
\begin{table}[!t]
\centering
\small
 \caption{Ablation study of BGA on the PASCAL-S, HKU-IS and DUTS-TE datasets, where `B' is the baseline model, and `G' denotes the guidance stage, and the `A' represents the aggregation stage.}
 \label{as-2}
 \setlength{\tabcolsep}{0.6mm}{
\begin{tabular}{ccc|cc|cc|cc}
\hline
\multirow{2}{*}{B}  & \multirow{2}{*}{\makecell[c]{G}} & \multirow{2}{*}{\makecell[c]{A}} & \multicolumn{2}{c|}{PASCAL-S}& \multicolumn{2}{c|}{DUTS-TE} & \multicolumn{2}{c}{HKU-IS} \\ \cline{4-9}
 &  &  &${{F}_{\beta}^{max}}\uparrow$ &${{MAE}}\downarrow$ &${{F}_{\beta}^{max}}\uparrow$ &${{MAE}}\downarrow$  &${{F}_{\beta}^{max}}\uparrow$ &${{MAE}}\downarrow$          \\ \hline
                \checkmark &           &            & 0.792         & 0.098         & 0.766          & 0.069         & 0.865         & 0.053        \\
                \checkmark &\checkmark &            & 0.803         & 0.086         & 0.786          & 0.056         & 0.880         &0.045         \\
                \checkmark &           &\checkmark  & 0.809         & 0.083         & 0.791          & 0.053         & 0.882         & 0.043         \\
                \checkmark &\checkmark & \checkmark &0.827            & 0.076           & 0.803     &  0.050   &0.892 & 0.038         \\ \hline
\end{tabular}}
\end{table}

\begin{figure}[!t]
	\centering
	\includegraphics[scale=0.35]{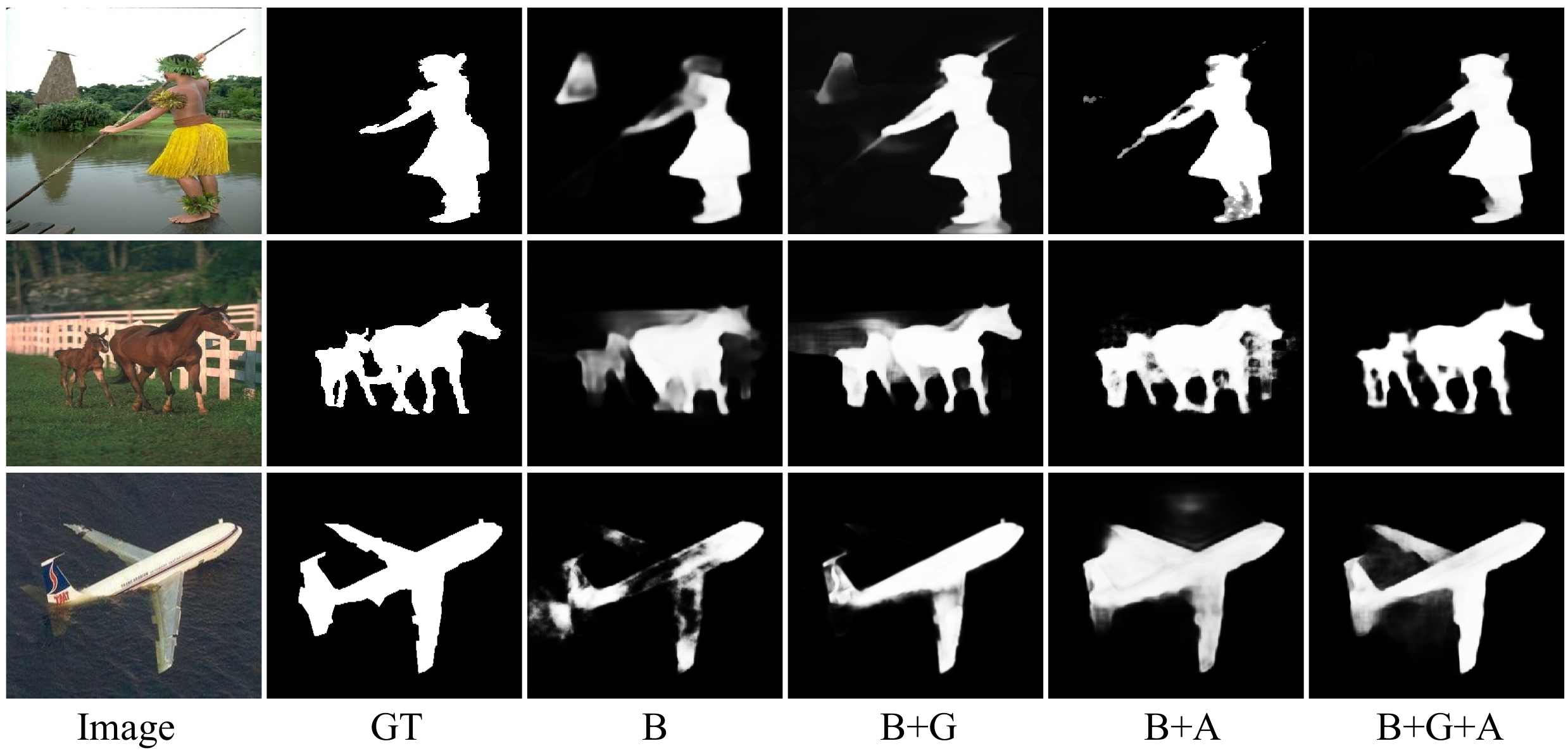}
	\caption{Visual comparisons for showing the benefits of the proposed modules.}
	\label{visualization-ab2}
\end{figure}
\subsubsection{Effectiveness of the R-Net}
In the design of R-Net, the BGA module is a crucial core module, including the guidance stage and aggregation stage.
In order to demonstrate the effectiveness of the designed BGA module R-Net, we conduct the ablation experiments on the PASCAL-S, DUTS-TE and HKU-IS datasets, and the quantitative and quantitative results are reported in Table \ref{as-2} and Fig. \ref{visualization-ab2}.

First, we replace the BGA module with a simple concatenation-convolution fusion, thereby forming the baseline model.
Based on the baseline model, we separately add the guidance and aggregation stages in the verification experiments.
In the guidance stage, the saliency-refinement mainstream branch is supplemented and enriched with the guidance information (\eg, object localization and completeness) of the RGB-image guidance branch. As shown in the Table \ref{as-2}, compared with the baseline model on the DUTS-TE dataset, the max F-measure is improved from 0.766 to 0.786 by only introducing the guidance stage, with a percentage gain of 2.6\%. The visualization results in Fig. \ref{visualization-ab2} show that some irrelevant backgrounds are effectively suppressed (such as the left region in the first image), and some salient regions can also be recovered (such as the lower wing region in the third image), but there are also some cases where the detection is incomplete (such as top wing in the third image).
In addition, the aggregation stage aims to more comprehensively integrate the corresponding encoder features, previous decoder features, and global context features.
When only the aggregation stage is introduced, we can achieve better performance than the baseline model, and even slightly better than the model only with the guidance stage, which also illustrates the importance of effective multi-level fusion. For example, only with the aggregation stage on the DUTS-TE dataset, the max F-measure is improved from 0.766 to 0.791, with a percentage gain of 3.3\%. From the visualization results, it can be seen that the aggregation stage can better complete the object structure (such as the upper wing in the third image), but still introduces some additional noise and interference.
By contrast, the model that includes both stages achieves the best performance. On the PASCAL-S, DUTS-TE and HKU-IS datasets, the percentage gain of the max F-measure reaches 4.4\%, 4.8\% and 3.1\% against the baseline model, respectively.
Also, the structure of the final result is more complete, and irrelevant background regions are suppressed more thoroughly.

Besides, to verify the effectiveness of the introduction of the RGB-image guidance branch in the R-Net, we add an ablation experiment. As a comparison, we remove the RGB-image guidance branch from the full model, denoted as w/o RGB. From the Table \ref{rm-rgb_branch}, we can see that the performance of the network degrades after removing the RGB branch on three testing datasets. For example, on the DUTS-TE dataset, compared with the model without the RGB branch, the max F-measure score is improved from 0.761 to 0.803 with a percentage gain of 5.5\%, and the MAE score is improved from 0.077 to 0.050 with a percentage gain of 35.1\%.

\begin{table}[!t]
\centering
\renewcommand\arraystretch{1.4}
    \begin{spacing}{1}
    \caption{Ablation study of RGB Branch in R-Net on the PASCAL-S, DUTS-TE and HKU-IS datasets.}
    \label{rm-rgb_branch}
    \setlength{\tabcolsep}{0.15mm}{
        \begin{tabular}{c|cc|cc|cc}
        \hline
                      & \multicolumn{2}{c|}{PASCAL-S} & \multicolumn{2}{c|}{DUTS-TE}
                      & \multicolumn{2}{c}{HKU-IS}\\ \cline{2-7}
                     &  ${{F}_{\beta }^{max}}\uparrow$      & $MAE\downarrow $
                     &   ${{F}_{\beta }^{max}}\uparrow$      & $MAE\downarrow $
                     &  ${{F}_{\beta }^{max}}\uparrow$      & $MAE\downarrow $  \\ \hline
       \makecell[c]{w/o RGB } & 0.791            & 0.109           & 0.761           & 0.077             & 0.873         &  0.054           \\
        Full model     &0.827    & 0.076  & 0.803  & 0.050   &0.892 & 0.038            \\ \hline
        \end{tabular}}
    \end{spacing}
\end{table}
\subsubsection{Effectiveness of Training Strategy}

To validate the effectiveness of our proposed training strategy, we conduct two ablation experiments:
(1) No.1: we simplify the designed training strategy. The training of the two networks is no longer performed alternately, but directly trains R-Net on 1,000 real-labeled samples, then tests 9,000 samples with coarse labels to obtain the corresponding pseudo labels, and finally uses the pseudo-labeled and real-labeled samples to train the S-Net. This experiment is designed to verify the effectiveness of the overall training strategy.
(2) No.2: we remove the credibility verification mechanism in each iteration to verify its effectiveness.
(3) No.3: we use all refined coarse labels and real labels as supervision in the fifth iteration of S-Net training.
(4) No.4: we remove contaminated data from each iteration and participate in each iteration using real-labeled samples, which is used to verify the effectiveness of the contamination mechanism on the real-labeled data.

As can be found in Table \ref{as-3}, even with our designed framework pipeline of label refinement and SOD, without our proposed training strategy, the network cannot exert its maximum advantage. For example, on the DUTS-TE dataset, the max F-measure of experiment No.1 drops from 0.803 to 0.763 compared to the model with full training strategy, a decrease of 4\%.
In addition, we introduce a credibility verification mechanism from the second iteration to ensure the validity of the pseudo labels. It can also be seen from the table that after removing this mechanism for experiment No.2, the indicators on all datasets decreased.
 From experiment No.3, we can see that although the amount of data for training S-Net is increased, the balance of data is disrupted, resulting in a slight decrease in the performance of the model instead of increasing. For example, compared with original training strategy, the max F-measure score of experiment No.3 drops from 0.827 to 0.824 on PASCAL-S dataset, and from 0.803 to 0.800 on the DUTS-TE dataset. Moreover, the training time of experiment No.3 is much longer than the full model.
 In addition, the role of the contamination labels is to prevent the network from overfitting the 1,000 samples with real labels, since these samples are involved in each round of training. From Table \ref{as-3}, it can be seen that the participation of contaminated real-labeled data in training improves the robustness and performance of the network. Compared to the version without the contamination mechanism on the real-labeled data, the MAE score is improved from 0.080 to 0.076 with a percentage gain of 5.2\% on the PASCAL-S dataset, and from 0.042 to 0.038 with a percentage gain of 10.5\% on the HKU-IS dataset.
In summary, our model framework equipped with the designed training strategy achieves apparent advantages in the detection performance on the PASCAL-S, HKU-IS and DUTS-TE datasets.

\renewcommand\arraystretch{1.1}
\begin{table}[!t]
\centering
    \caption{Ablation study of Training Strategy with Hybrid Labels on PASCAL-S, DUTS-TE and HKU-IS datasets.}
    \label{as-3}
    \setlength{\tabcolsep}{0.6mm}{
        \begin{tabular}{c|cc|cc|cc}
        \hline

                      & \multicolumn{2}{c|}{PASCAL-S} & \multicolumn{2}{c|}{DUTS-TE}
                      & \multicolumn{2}{c}{HKU-IS}\\ \cline{2-7}
                     & ${{F}_{\beta }^{max}}\uparrow$      & $MAE\downarrow $
                     & ${{F}_{\beta }^{max}}\uparrow$      & $MAE\downarrow $
                     & ${{F}_{\beta }^{max}}\uparrow$      & $MAE\downarrow $
                     \\ \hline
       No.1          & 0.786             & 0.095           &0.763      & 0.071  &0.863 &  0.059             \\
       No.2          & 0.819             & 0.083           & 0.792     &0.054   &0.880 &  0.043             \\
       No.3  & 0.824  & 0.075 & 0.800    &0.050   &0.889 &  0.039             \\\hline
       \makecell[c]{No.4} & 0.812  & 0.080 & 0.796 & 0.053  & 0.878 &  0.042     \\\hline
        Ours        &0.827            & 0.076           & 0.803     &  0.050   &0.892 & 0.038            \\ \hline
        \end{tabular}}

\end{table}


\subsubsection{Impact of the Group Settings}
Considering the sample imbalance issue caused by the difference in the number of real-labeled samples and coarse-labeled samples, we propose a group-wise incremental mechanism to avoid network collapse caused by importing a large amount of pseudo-labeled data at a time. In implementation, we divide all training samples into some groups, and gradually increase the amount of data with pseudo labels in each training iteration. The number of groups simply reflects the number of samples embedded in each iteration and does not have a significant impact on performance theoretically, but the larger the number of groups, the more iterations required and the longer the training time. To this end, we design ablation experiments with different number of groups (\ie, 5 and 15), as reported in Table \ref{group-settings}. When the number of groups is set to 5, the 1,000 samples with real labels are still grouped into GROUP 1. The remaining 9,000 samples with coarse labels are divided into four groups for training, each containing 2,250 samples. Since the number of training iterations is related to the number of groups, we add one group of data for each iteration, and these five groups are iterated three times in total. Similarly, when the number of groups is set to 15, the remaining 9,000 samples with coarse labels are divided into 14 groups of 642 or 643 images each, and the whole training process requires 8 iterations. It can be seen from Table \ref{group-settings} that the performance of different grouping numbers is slightly different, which is also consistent with our theoretical analysis.

\begin{table}[!t]
\centering
\small
\renewcommand\arraystretch{1.5}
    \begin{spacing}{1}
    \caption{Ablation study of Different Group Settings.}
    \label{group-settings}
    \setlength{\tabcolsep}{0.1mm}{
        \begin{tabular}{c|cc|cc|cc}
        \hline
                      & \multicolumn{2}{c|}{PASCAL-S} & \multicolumn{2}{c|}{DUTS-TE}
                      & \multicolumn{2}{c}{HKU-IS}\\ \cline{2-7}
                     & ${{F}_{\beta }^{max}}\uparrow$      & $MAE\downarrow $
                     & ${{F}_{\beta }^{max}}\uparrow$      & $MAE\downarrow $
                     & ${{F}_{\beta }^{max}}\uparrow$      & $MAE\downarrow $  \\ \hline
        5 Groups & 0.824  &0.074   & 0.798  & 0.049  & 0.889      &  0.039           \\
        15 Groups& 0.824  & 0.075  &  0.801 & 0.049  & 0.890      &  0.039           \\
        \makecell[c]{10 Groups (Ours) }    &0.827     & 0.076  & 0.803     &  0.050  &0.892 & 0.038            \\ \hline
        \end{tabular}}
    \end{spacing}
\end{table}

\section{Conclusion}
In this paper, we propose a weakly-supervised learning framework for SOD tasks with hybrid labels, which is decoupled into a R-Net and a S-Net. In order to make full use of the limited annotation information, the R-Net equipped with Blender with Guidance and Aggregation Mechanisms is designed to refine the coarse label and generate the pseudo label for S-Net training.
In addition, we design three training mechanisms to guarantee the effectiveness and efficiency of network training, including alternate iteration mechanism, group-wise incremental mechanism, and credibility verification mechanism.
Evaluations of five benchmark datasets demonstrate the effectiveness of our approach.


%




\ifCLASSOPTIONcaptionsoff
  \newpage
\fi



%
\bibliographystyle{IEEEtran}
\bibliography{sample-base}

\end{document}